# Asymptotic Model Selection for Naive Bayesian Networks

Dmitry Rusakov and Dan Geiger
Computer Science Department
Technion - Israel Institute of Technology
Haifa, Israel 32000
{rusakov,dang}@cs.technion.ac.il

## Abstract

We develop a closed form asymptotic formula to compute the marginal likelihood of data given a naive Bayesian network model with two hidden states and binary features. This formula deviates from the standard BIC score. Our work provides a concrete example that the BIC score is generally not valid for statistical models that belong to a stratified exponential family. This stands in contrast to linear and curved exponential families, where the BIC score has been proven to provide a correct approximation for the marginal likelihood.

## 1 INTRODUCTION

Statisticians are often faced with the problem of choosing the appropriate model that best fits a given set of observations. One example of such problem is the choice of structure in learning of Bayesian networks (Heckerman, Geiger & Chickering, 1995; Cooper & Herskovits, 1992). In such cases the maximum likelihood principle would tend to select the model of highest possible dimension, contrary to the intuitive notion of choosing the right model. Penalized likelihood approaches such as AIC have been proposed to remedy this deficiency (Akaike, 1974).

We focus on the Bayesian approach to model selection, by which a model $M$ is chosen according to the maximum posteriori probability given the observed data $D$:

$$P(M|D) \propto P(M,D) = P(M)P(D|M)$$
$$= P(M) \int_\Omega P(D|M,\omega)P(\omega|M)d\omega$$

where $\omega$ denotes the model parameters and $\Omega$ denotes the domain of the model parameters. In particular we focus on large sample approximation for $P(M|D)$.

The critical computational part in the evaluation of this criterion is the marginal likelihood integral $\mathbb{I} \equiv P(D|M) = \int_\Omega P(D|M,\omega)P(\omega|M)d\omega$. We write

$$\mathbb{I}[N,Y_D,M] = \int_\Omega e^{loglikelihood(Y_D,N|\omega,M)}\mu(\omega|M)d\omega \quad (1)$$

where $Y_D$ is the averaged sufficient statistics of the data $D$, $N$ is a number of examples in $D$, and $\mu(\omega|M)$ is the prior parameter density for model $M$. Recall that the average sufficient statistics for multinomial samples of $n$ binary variables $(X_1,\ldots,X_n)$ is simply the counts for each of the possible $2^n$ joint states. Often the prior $P(M)$ is assumed to be equal for all models, in which case Bayesian model selection is performed by maximizing $\mathbb{I}[N,Y_D,M]$. The quantity represented by $S(Y_D,N,M) \equiv \ln \mathbb{I}[N,Y_D,M]$ is called the *Bayesian Information Criterion (BIC)* for choosing model $M$.

For many types of models the asymptotic evaluation of integral 1 (as $N \to \infty$) is a classical Laplace procedure. This evaluation was first performed for Linear Exponential (LE) models (Schwarz, 1978) and then for Curved Exponential (CE) models under some additional technical assumptions (Haughton, 1988). It was shown that

$$S(Y_D,N,M) = N \cdot \ln P(Y_D|\omega_{ML}) - \frac{d}{2}\ln N + R, \quad (2)$$

where $\ln P(Y_D|\omega_{ML})$ is the log-likelihood of $Y_D$ given the maximum likelihood parameters of the model and $d$ is the model dimension, i.e., the number of independent parameters. The error term $R = R(Y_D,N,M)$ was shown to be bounded for a fixed $Y_D$ (Schwarz, 1978) and uniformly bounded for all $Y_D \to Y$ in CE models (Haughton, 1988). This approximation is referred as a (standard) *BIC score*.

The use of BIC score for Bayesian model selection for Graphical Models is valid for Undirected Graphical Models without hidden variables because these are LE models (Lauritzen, 1996). The justification of BIC for Directed Graphical Models (called Bayesian Networks) is somewhat more complicated. On one hand discrete and Gaussian DAG models are CE models (Geiger, Heckerman, King & Meek, 2001; Spirtes, Richardson & Meek, 1997). On the other hand, the theoretical justification of the BIC score for CE models has been established under the assumption that the model contains the true distribution - the one that has generated the observed data. This assumption limits the applicability of the proof of BIC score's validity for Bayesian networks in practical setups.

The evaluation of the marginal likelihood $\mathbb{I}[N,Y]$ for Bayesian networks with hidden variables is a wide open problem because the class of distributions represented by Bayesian networks with hidden variables is significantly richer than curved exponential models and it falls into the class of Stratified Exponential (SE) models (Geiger et al., 2001). For such models the effective dimensionality $d$ (Eq. 2) of the model is no longer the number of network parameters (Geiger, Heckerman & Meek, 1996; Settimi & Smith, 1998). Moreover, the central problem in the evaluation of the marginal likelihood for this class is that the set of maximum likelihood points is sometimes a complex self-crossing surface. Recently, major progress has been achieved in analyzing and evaluating this type of integrals (Watanabe, 2001). Herein, we apply these techniques to model selection among Bayesian networks with hidden variables.

The focus of this paper is the asymptotic evaluation of $\mathbb{I}[N,Y,M]$ for a binary naive Bayesian model $M$ with binary features. The results are derived under similar assumptions to the ones made by Schwarz (1978) and Haughton (1988). In this sense, our paper generalizes the mentioned works, providing valid asymptotic formulas for a new type of marginal likelihood integrals. The resulting asymptotic approximations, presented in Theorem 3, deviate from the standard BIC score. Hence the standard BIC score is not justified for Bayesian model selection among Bayesian networks with hidden variables. Our *adjusted* BIC score changes depending on the different types of singularities of the sufficient statistics, namely, the coefficient of the $\ln N$ term is no longer $-\frac{d}{2}$ but rather a function of the sufficient statistics. Moreover, an additional $\ln \ln N$ term appears in some of the $O(1)$ approximations, which is unaccounted for by the classical score.

The rest of this paper is organized as follows. Section 2 introduces the concept of asymptotic expansions and presents some methods of asymptotic approximation. Section 3 discusses an application of these methods. Section 4 reviews naive Bayesian models and explicates the relevant marginal likelihood integrals for these models. Section 5 states and explains our main result deferring the proof to an Appendix. Finally, Section 6 outlines future research directions.

## 2 ASYMPTOTIC APPROXIMATIONS

Exact analytical formulas are not available for many integrals arising in practice. In such cases some sort of approximate or asymptotic solutions are of interest. Asymptotic analysis is a branch of analysis that is concerned with obtaining the approximate analytical solutions to problems where a parameter or some variable in an equation or integral becomes either very large or very small.

Let $z$ represent such a large parameter. We say that $f(z)$ is *asymptotically equal* to $\sum_{n=1}^{m} a_n g_n(z)$, denoted by the symbol "$\sim$", if

$$f(z) = \sum_{n=1}^{m} a_n g_n(z) + O(g_{m+1}(z)), \text{ as } z \to \infty,$$

where the big $O$ symbol states that the error term is bounded by a constant multiply of $g_{m+1}(z)$ and $\{g_n\}$ is an *asymptotic sequence*, i.e., $\lim_{z \to \infty} g_{n+1}/g_n = 0$. A good introduction to asymptotic analysis can be found in (Murray, 1984).

The main objective of this paper is asymptotic approximation of marginal likelihood integrals as represented by Eq. 1, which are of the form

$$\mathbb{I}[N,Y] = \int_\Omega e^{-Nf(\omega,Y)} \mu(\omega) d\omega \qquad (3)$$

where $f(\omega, Y) = -loglikelihood(Y|\omega)$. We shall assume that we are dealing with exponential models, so the log-likelihood of sampled data is equal to $N$ times the log-likelihood of the averaged sufficient statistics. This assumption holds for the models discussed in this paper.

Consider Eq. 3 for some fixed $Y$. For large $N$, the main contribution to the integral comes from the neighborhood of the minimum of $f$, i.e., the maximum of $-Nf(\omega, Y)$. Thus, intuitively, the approximation of $\mathbb{I}[N,Y]$ is determined by the form of $f$ near its minimum on $\Omega$. In the simplest case $f(\omega)$ achieves a single minimum at $\omega_{ML}$ in the interior of $\Omega$ and this maximum is non-degenerate, i.e., the Hessian matrix $\mathcal{H}f(\omega_{ML})$ of $f$ at $\omega_{ML}$ is of full rank. In this case the approximation of $\mathbb{I}[N,Y]$ for $N \to \infty$ is the classical Laplace procedure (e.g., Wong, 1989, page 495), summarized as follows

**Lemma 1 (Laplace Approximation)** *Let*

$$I(N) = \int_U e^{-Nf(u)} \mu(u) du$$

where $U \subset \mathbb{R}^d$. Suppose that $f$ is twice differentiable and convex ($\mathcal{H}f(u) > 0$), the minimum of $f$ on $U$ is achieved on a single internal point $u_0$, $\mu$ is continuous and $\mu(u_0) \neq 0$. If $I(N)$ absolutely converges, then

$$I(N) \sim C e^{-Nf(u_0)} N^{-d/2} \qquad (4)$$

where $C = (2\pi)^{d/2} \mu(u_0) [\det \mathcal{H}f(u_0)]^{-\frac{1}{2}}$ is a constant.

Note that the logarithm of Eq. 4 yields the BIC score as presented by Eq. 2.

However, in many cases, and, in particular, in the case of naive Bayesian networks, the minimum of $f$ is achieved not at a single point in $\Omega$ but rather on a variety $W_0 \subset \Omega$, called the zero set. Sometimes, this variety may be $d'$-dimensional surface (smooth manifold) in $\Omega$ in which case the calculation of the integral is locally equivalent to the $d - d'$ dimensional classical case. The hardest cases to evaluate happen when the variety $W_0$ contains self-crossings.

Fortunately, an advanced mathematical method for approximating this type of integrals was introduced to the machine learning community by Watanabe (2001). Below we introduce the main theorem that enables us to compute the asymptotic form of $\mathbb{I}[N, Y]$ integrated in a neighborhood of a maximum likelihood point.

**Theorem 2 (based on (Watanabe, 2001))** *Let*

$$I(N) = \int_{W_\epsilon} e^{-Nf(w)} \mu(w) dw$$

*where $W_\epsilon$ is some closed $\varepsilon$-box around $w_0$, which is a minimum point of $f$ in $W_\epsilon$, and $f(w_0) = 0$. Assume that $f$ and $\mu$ are analytic functions, $\mu(w_0) \neq 0$. Then,*

$$\ln I(N) = \lambda_1 \ln N + (m_1 - 1) \ln \ln N + O(1) \qquad (5)$$

*where the rational number $\lambda_1 < 0$ and the natural number $m_1$ are the largest pole and its multiplicity of the meromorphic (analytic + poles) function that is analytically continued from*

$$J(\lambda) = \int_{f(w) < \epsilon} f(w)^\lambda \mu(w) dw \qquad (Re(\lambda) > 0)$$

*where $\epsilon > 0$ is a sufficiently small constant.*

Applying Theorem 2 to the classical, single maximum, strictly convex case (Lemma 1) gives the largest pole $\lambda_1 = -d/2$, with multiplicity $m = 1$ confirming the classical result (Example 1 in (Watanabe, 2001)). However in the more complex cases, e.g., when the integral is evaluated in the neighborhood of the self-crossing of the zero set $W_0$, the coefficient $-\lambda_1$ is not equal to half the dimensionality of the parameter space. In fact, $2\lambda_1$ need not be an integer; it is a rational number.

In general it is not easy to find the largest pole and multiplicity of $J(\lambda)$. Here, another fundamental mathematical theory comes to rescue. The resolution of singularities in algebraic geometry transforms the integral $J(\lambda)$ into a direct product of integrals of a single variable (Atiyah, 1970, Resolution Theorem; Hironaka, 1964). We demonstrate this technique in the next section.

## 3 APPLICATION OF WATANABE'S METHOD

We now apply the method of Watanabe (2001) to approximate the integral

$$\tilde{\mathbb{J}}[N] = \int_{(-\epsilon, +\epsilon)^n} e^{-N \sum_{1 \leq l \neq k \leq n} u_l^2 u_k^2} du$$

as $N$ tends to infinity. This evaluation is part of the proof of our main result (Theorem 3). It is presented here as a self contained example which can be skipped without loss of continuity.

Watanabe's method calls for the analysis of the poles of the following function

$$J(\lambda) = \int_{(-\epsilon, +\epsilon)^n} \left[ \sum_{1 \leq l \neq k \leq n} u_l^2 u_k^2 \right]^\lambda du.$$

To find the poles we transform the integrand function $\psi(u) = \sum_{1 \leq l \neq k \leq n} u_l^2 u_k^2$ into a function of new coordinates $v_1, \ldots, v_n$ such that $\psi(v) = a(v) v_1^{\alpha_1} v_2^{\alpha_2} \ldots v_n^{\alpha_n}$ and $a(v)$ is invertible near 0. This transformation decompose the integral under study into $n$ independent one-dimensional integrals each of which can be easily computed. The process of changing to the new coordinates is known as the process of *resolution of singularities*. To obtain the needed transformations for the integral under study, we apply a technique called *blowing-up* which consists of a series of *quadratic transformations*. For an accessible introduction to these concepts see (Abhyankar, 1990).

We start with $n = 3$ and then generalize. Rescaling the integration range to $(-1, 1)$ and then taking only the positive quadrant, which does not change the poles of $J(\lambda)$, yields

$$\begin{aligned}
J(\lambda) &= \int_{(0,1)^3} (u_1^2 u_2^2 + u_1^2 u_3^2 + u_2^2 u_3^2)^\lambda du \\
&= \left( \int_{0 < u_2, u_3 < u_1 < 1} + \int_{0 < u_1, u_3 < u_2 < 1} + \int_{0 < u_1, u_2 < u_3 < 1} \right) (u_1^2 u_2^2 + u_1^2 u_3^2 + u_2^2 u_3^2)^\lambda du.
\end{aligned}$$

The three cases are symmetric, so we evaluate only the first. Using the quadratic transformation $u_2 = u_1 u_2$, $u_3 = u_1 u_3$, yields

$$\begin{aligned}
J_1(\lambda) &= \int_{0 < u_2, u_3 < u_1 < 1} (u_1^2 u_2^2 + u_1^2 u_3^2 + u_2^2 u_3^2)^\lambda du \\
&= \int_{(0,1)^3} u_1^{4\lambda+2} (u_2^2 + u_3^2 + u_2^2 u_3^2)^\lambda du.
\end{aligned}$$

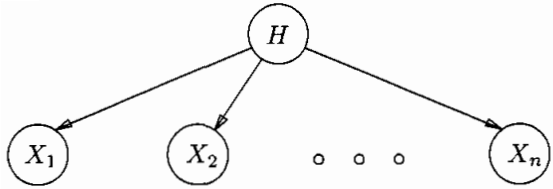

Figure 1. A naive Bayesian model.

We now divide the range $(0,1)^3$ according to $u_2 < u_3$ or $u_3 < u_2$. Again these cases are symmetric and so we continue to evaluate only one of them

$$J_{11}(\lambda) = \int_{u_3<u_2} u_1^{4\lambda+2}(u_2^2+u_3^2+u_2^2u_3^2)^\lambda du$$
$$= \int_{(0,1)^3} u_1^{4\lambda+2} u_2^{2\lambda+1}(1+u_3^2+u_2^2u_3^2) du.$$

Since the function $(1+u_3^2+u_2^2u_3^2)$ is bounded on $(0,1)^3$, it follows that $J(\lambda)$ is within a constant multiply of

$$J(\lambda) \approx \int_{(0,1)^2} u_1^{4\lambda+2} u_2^{2\lambda+1} du.$$

Thus $J(\lambda)$ has poles at $\lambda = -3/4$ and $\lambda = -1$ with multiplicity $m = 1$. The largest pole is $\lambda = -3/4$ with multiplicity $m = 1$. Generalizing the above approach to $n \geq 3$ we get that the largest pole of $J(\lambda)$ is $\lambda_1 = -n/4$ with multiplicity $m = 1$, so $\mathbb{J}[N]$ is asymptotically equal to $cN^{-\frac{n}{4}}$.

## 4 NAIVE BAYESIAN MODELS

A naive Bayesian model $M$ for discrete variables $X = \{X_1, \ldots, X_n\}$ is a set of joint distributions for $X$ that factor according to the tree structure depicted on Figure 1. A probability distribution $P(x)$ belongs to a naive Bayesian model if

$$P(x) = \sum_{j=1}^r P(H = h_j) \prod_{i=1}^n P(X_i = x_i | H = h_j),$$

where $x$ is the $n$-dimensional binary vector of values of $X$, $r$ is the number of hidden states and $h_j$ denotes a particular state (class). Intuitively, this model describes the generation of data $x$ that comes from $r$ sources $h_1, \ldots, h_r$. Naive Bayesian models are a subclass of Bayesian networks (Pearl, 1988).

In this work we focus on naive Bayesian networks that have two hidden states ($r = 2$) and $n$ binary feature variables $X_1, \ldots, X_n$. We denote the parameters defining $p(x_i|c_1)$ by $a_i$, the parameters defining $p(x_i|c_2)$ by $b_i$, and the parameters defining $p(c_1)$ by $t$. These parameters are called the *model parameters*. We denote the *joint space parameters* $P(X = x)$ by $\theta_x$. The following mapping relates these parameters.

$$\theta_x = t \prod_{i=1}^n a_i^{x_i}(1-a_i)^{1-x_i} + (1-t)\prod_{i=1}^n b_i^{x_i}(1-b_i)^{1-x_i}, \quad (6)$$

and the marginal likelihood integral (1) becomes

$$\mathbb{I}[N,Y] = \int_{(0,1)^{2n+1}} e^{N \sum_x Y_x \ln \theta_x(\omega)} \mu(\omega) d\omega \quad (7)$$

where $\omega = (a_1, \ldots, a_n, b_1, \ldots, b_n, t)$ are the model parameters.

## 5 MAIN RESULT

This section presents an asymptotic approximation of the integral $\mathbb{I}[N,Y]$ (Eq. 7) for naive Bayesian networks consisting of binary variables $X_1, \ldots, X_n$ and two hidden states. It is based on two results. First, the classification of singular points for these types of models (Geiger et al., 2001). Second, Watanabe's approach as explained in Section 2, which provides a method to obtain the correct asymptotic formula of $\mathbb{I}[N,Y]$ for the singular points not covered by the classical Laplace approximation scheme.

Let $\Upsilon = \{(y_1, \ldots, y_{2^n}) | y_i \geq 0, \sum y_i = 1\}$ be the set of possible values of sufficient statistics $Y = (Y_1, \ldots, Y_{2^n})$ for data $D = \{(x_{i,1}, \ldots, x_{i,n})\}_{i=1}^N$. In our asymptotic analysis we let the sample size $N$ grow to infinity.

Let $\Upsilon_0 \subset \Upsilon$ be the points $(y_1, \ldots, y_{2^n})$ that correspond to the distributions that can be represented by binary naive Bayesian models with $n$ binary variables. I.e., assuming the indices of $y_i$ are written as vectors $(\delta_1, \ldots, \delta_n)$ of $n$ zeros and ones, points in $S$ are those that can be parameterized via

$$y_{(\delta_1,\ldots,\delta_n)} = t \prod a_i^{\delta_i}(1-a_i)^{1-\delta_i} + \quad (8)$$
$$(1-t)\prod b_i^{\delta_i}(1-b_i)^{1-\delta_i}$$

where $t$, $a = (a_1, \ldots, a_n)$ and $b = (b_1, \ldots, b_n)$ are the $2n+1$ model parameters, as defined in Section 4.

Geiger et al. (2001) classify the singular points into two classes $S$ and $S'$. The set $S$ is the set of points $(y_1, \ldots, y_{2^n})$ such that Eq. 8 holds and all $a_i = b_i$ except for at most two indices in $\{1, \ldots, n\}$. Intuitively, each such point represents a probability distribution that can be defined by a naive Bayesian model (Figure 1) with all links removed except at most two.

The set $S' \subset S$ is the set of points represented by a naive Bayesian model, just as the set $S$ does, but with all links removed; namely, a distribution where all variables are mutually independent and independent of the class node as well.

Clearly $S' \subset S \subset \Upsilon_0 \subset \Upsilon$. We now present our main result.

**Theorem 3** *Let $\mathbb{I}[N,Y]$ be the marginal likelihood of data with sufficient statistics $Y$ given the naive*

*Bayesian model with binary variables and two hidden states, as represented by Eqs. 6 and 7. Let $Y$ and $\mu$ satisfy following assumptions:*

**A1** Bounded density. *The density $\mu(\omega)$ is bounded and bounded away from zero on $\Omega = (0,1)^{2n+1}$.*

**A2** Positive statistics. *The statistics $Y$ are such that $Y_i > 0$ for $i = 1, \ldots, 2^n$.*

**A3** Statistics stability. *There exists sample size $N_0$ such that the sufficient statistics is $Y$ for all sample sizes $N > N_0$.*

*Then, for $n \geq 3$ as $N \to \infty$:*

1. *If $Y \in \Upsilon_0 \setminus S$ (regular point)*

$$\ln \mathbb{I}[N, Y] = N f_Y - \frac{2n+1}{2} \ln N + O(1), \quad (9)$$

2. *If $Y \in S \setminus S'$ (type 1 singularity)*

$$\ln \mathbb{I}[N, Y] = N f_Y - \frac{2n-1}{2} \ln N + O(1), \quad (10)$$

3. *If $Y \in S'$ (type 2 singularity)*

$$\ln \mathbb{I}[N, Y] = N f_Y - \frac{n+1}{2} \ln N + O(1), \quad (11)$$

*where $f_Y = \ln P(Y|\omega_{ML})$ and $\omega_{ML}$ is the maximum likelihood parameters.*
*Moreover, for $n = 1, 2$ (degenerate models),*

- *If $n = 2$, and $Y \notin S'$, or $n = 1$, then*

$$\ln \mathbb{I}[N, Y] = N f_Y - \frac{2n-1}{2} \ln N + O(1), \quad (12)$$

- *If $n = 2$ and $Y \in S'$,*

$$\ln \mathbb{I}[N, Y] = N f_Y - \frac{3}{2} \ln N + 2 \ln \ln N + O(1), \quad (13)$$

*as $N \to \infty$.*

The first assumption (bounded density) has been made by all earlier works; in some applications they hold and in some they do not. The proof and the results, however, can be easily modified to apply to any particular kind of singularity of $\mu$, as long as we know the form of this singularity. The second and third assumptions are made to ease the proof; the third assumption was also made by Schwarz (1978).

Note that Eq. 10 corresponds to selecting $\lambda_1 = -\frac{2n-1}{2}$ and $m_1 = 1$ in Watanabe's method, Eq. 11 corresponds to selecting $\lambda_1 = -\frac{n+1}{2}$ and $m_1 = 1$, and Eq. 13 corresponds to selecting $\lambda_1 = -\frac{3}{2}$ and $m = 3$. These formulas are different from the standard BIC score, given by Eq. 9, which only applies to regular points in non-degenerate models, namely, the points in $\Upsilon_0 \setminus S$.

In contrast to the standard BIC score, which is uniform for all points $Y$, the asymptotic approximation given by our *adjusted BIC score* depends on the value of $Y$ through the coefficient of $\ln N$. This coefficient in the singular cases is not the effective dimensionality of the parameter space, because the parametric space is singular at these points, namely, not isomorphic to any hypersurface. Instead, the coefficients of $\ln N$ and $\ln \ln N$ terms describe the geometric structure of the log-likelihood function near singular points. E.g., for $n = 2$ and $Y \in S'$, we get the $2 \ln \ln N$ term in $O(1)$ approximation, which is missed by the standard BIC score formula, Eq. 2.

One may argue that evaluating the marginal likelihood on singular points is not needed because one could exclude from the model all singular points which only have measure zero (with respect to the volume element of the highest dimension). The remaining set would be a smooth manifold defining a curved exponential model, and so BIC would be a correct asymptotic expansion as long as the point $Y$ has not been excluded, i.e., it will be correct for regular $Y \in \Upsilon_0$ points. However, this proposal fails in the case of selecting amongst naive Bayesian models.

Consider the problem of selecting between two naive Bayesian models with $n$ binary features and a binary hidden class variable. One model $M_F$ with all links present between the class variable and each and every feature variable, and the other being a degenerate model $M_D$ for which all the feature variables are mutually independent and independent also of the class variable, namely, there are no links present. Assuming the two states $h_1$ and $h_2$ of the hidden variable are interpreted as representing two classes, model $M_D$ tells us that the $n$ features in the model do not distinguish between the two classes, and so if model $M_D$ is correct, the two classes are not distinguishable using the prescribed $n$ features. If model $M_F$ is correct, it provides some support for the existence of two classes, the strength of which is determined by the parameters of the model. Assume a prior probability of $p > 0$ and $1 - p > 0$ for the two models, respectively.

Now, if the true data comes from model $M_D$, then its large sample statistics falls very close to the set $S'$ of singular points of the full model $M_F$. Even if the statistics are regular due to small perturbations in the sample, evaluation of $\mathbb{I}[N, Y]$ according to the regular case formula will give very large error terms and result in an incorrect model selection. Hence, in this case, one should evaluate the marginal likelihood of $M_F$ using *uniform* asymptotic formulas, which are valid for the range of $Y$ near singular points, and which, in the limit, are equivalent to the formulas derived in this paper. This careful evaluation should be performed for a non negligible fraction $p$ of the possible large sample datasets, at least according to the prior specification. This phenomenon happens whenever comparing a graphical model against one of its submodels, which is a common practice that requires a careful analysis that this paper attempts to provide.

## 6 FUTURE WORK

We now highlight the steps required for obtaining a fully justified asymptotic model selection criterion for naive Bayesian networks.

1. Develop a closed form asymptotic formula for marginal likelihood integrals for all types of statistics $Y$ given an arbitrary naive Bayesian model. This step has been partially treated by the current paper.

2. Extend these solutions by developing *uniform* asymptotic formulas valid for converging statistics $Y_D \to Y$ as $N \to \infty$.

3. Develop an algorithm that, given a naive Bayesian network and a data set with statistics $Y_D$, determines the possible singularity types of the limit statistics $Y$ and applies the appropriate asymptotic formula developed in step 2.

Our work provides a first step and a concrete framework to resolve these tasks among naive Bayesian networks and perhaps among Bayesian networks with hidden variables in general.

**Acknowledgements**

The second author thanks David Heckerman and Chris Meek for years of collaboration on this subject.

# APPENDIX: PROOF OUTLINE

The integral $\mathbb{I}[N,Y]$ converges for all $N \geq 1$ and for all $Y$ because the likelihood function is bounded. The first claim of Theorem 3 follows from the fact that for $Y \in \Upsilon_0 \setminus S$ there are only two (symmetric) maximum likelihood points at each of which the log-likelihood function is properly convex. Hence, the marginal likelihood integral can be approximated by the classical Laplace method (Lemma 1).

The proof of the second and third claims of Theorem 3 requires the advanced techniques of Watanabe (Section 2). First, the integral $\mathbb{I}[N,Y]$ is transformed by a series of transformations into a simpler one. Second, the sets of extremum points of the exponent (maximum log-likelihood points) are found, and then the new integral is computed in the neighborhoods of extremum points. Finally, the asymptotic form of the largest contribution gives the desired asymptotic approximation to the original integral. We focus on one thread of our proof which demonstrates this method.

## USEFUL TRANSFORMATIONS

We first introduce a series of three transformations from the model parameters $\omega = (a, b, t)$ to the joint space parameters $\theta_x$ that facilitates the approximation of $\mathbb{I}[N,Y]$. The transformations $T_1$, $T_2$ and $T_3$ are such that their composition $T = T_3 \circ T_2 \circ T_1 : \Omega \to \Theta$ is defined by Eq. 6, where $\Omega = (0,1)^{2n+1}$ is the domain of model parameters $\omega$ and $\Theta$ is the domain of joint space parameters $\theta_x$. We call $\omega$'s - *the source variables* and $\theta$'s - *the target variables*. These transformations are from (Geiger et al., 2001).

**Transformation $T_1$:** Let $T_1 : \Omega \to U$ be defined via
$$s = 2t - 1, \quad u_i = \frac{a_i - b_i}{2}, \quad x_i = ta_i + (1-t)b_i,$$
$i = 1, \ldots, n$. The mapping $T_1$ is a diffeomorphism, namely, a one-to-one differentiable map with a differentiable inverse. Furthermore, $|\det J_{T_1}| = 2^{-n+1}$.

**Transformation $T_3$:** The transformation $T_3 : \Lambda \to \Theta$ is defined on the original target variables in such way that the new target variables $z \in \Lambda$ are expressed in terms of the new source variables $(x, u, s)$ by a number of simple formulas. The exact form of this transformation is unimportant for our analysis. We note that $T_3$ is diffeomorphism and $|\det J_{T_3}| = 1$. For details consult (Geiger et al., 2001).

**Transformation $T_2$:** This transformation is defined by $T_2 : U \subset \mathbb{R}^{2n+1} \to \Lambda \subset \mathbb{R}^{2^n-1}$ via
$$z_i = x_i, \quad z_{ij} = p_2(s)u_i u_j$$
$$\ldots$$
$$z_{12\ldots r} = p_r(s)u_1 u_2 \ldots u_r,$$

where $p_i(s) = \frac{1}{2}(1-s^2)((1-s)^{i-1} + (-1)^i(1+s)^{i-1})$, and, in particular, $p_2(s) = 1 - s^2$. We index the $z$ variables by non-empty subsets of $\{1,\ldots,n\}$. Note that, generally, this transformation is not a diffeomorphism.

We have defined three transformations, from the model parameters $\Omega$ to the joint space parameters $\Theta$:
$$\Omega \xleftrightarrow{T_1} U \xrightarrow{T_2} \Lambda \xleftrightarrow{T_3} \Theta$$

Based on these transformations we now present lemma that facilitates the evaluation of the integral $\mathbb{I}[N,Y]$.

**Lemma 4** *Let $\mathbb{I}[N,Y]$ be as represented by Eq. 7 and let $f$ define the normalized log-likelihood function, namely*
$$f(\theta) = f_Y - \sum_i Y_i \ln \theta_i,$$
*where $f_Y = \ln P(Y|\omega_{ML})$ and $\theta[x,u,s] = (T_3 \circ T_2)[x,u,s]$. Also, let the zero set $U_0 = \arg\min_{(x,u,s) \in \bar{U}} f(x,u,s)$ be the set of minimum points of $f$ in $(x,u,s)$ coordinates, and let*
$$\mathbb{J}[N,Y] = \sum_{(x,u,s)_i \in U_0} \int_{U_{\epsilon,i}} e^{-N \sum (z_I - z_{0i,I})^2} dx\, du\, ds, \quad (14)$$
*where $U_{i,\epsilon}$ is a small neighborhood of $(x,u,s)_i \in U_0$, $z(x,u,s) = T_2(x,u,s)$ and $z_{0i} = T_2[(x,u,s)_i]$. Suppose that $\mu$ is bounded (A1), and $Y$ is positive (A2) and fixed (A3). Then, for all $N > 1$,*
$$\ln \mathbb{I}[N,Y] = Nf_Y + \ln \mathbb{J}[N,Y] + O(1)$$
*where $\mathbb{J}[N,Y]$ is represented by finite sum of neighborhood integrals.*

This lemma follows from the assumptions A1-A3 and the facts that $T_1$, $T_3$ are diffeomorphisms, $Y \in \Upsilon_0$, $\bar{U}$ is compact, and the contributions of non-maximum regions of the integrand are exponentially small.

Lemma 4 states that integral 14 determines the asymptotic form of the original integral $\mathbb{I}[N,Y]$.

## PROOF FOR TYPE 2 SINGULARITY

We focus on the proof of the third claim of Theorem 3 that deals with the singular points in $S'$. This proof illustrates the methods required for the proofs of all cases of Theorem 3.

Let $Y \in S'$. Our starting point is integral 14, which by Lemma 4 is within a constant multiply of the original integral (without the $e^{-Nf_Y}$ term). We evaluate the contributions to the integral $\mathbb{J}[N,Y]$ from the neighborhoods of extremum points $(x', u', s') \in U_0$. The maximum contribution gives the asymptotic order of the original integral.

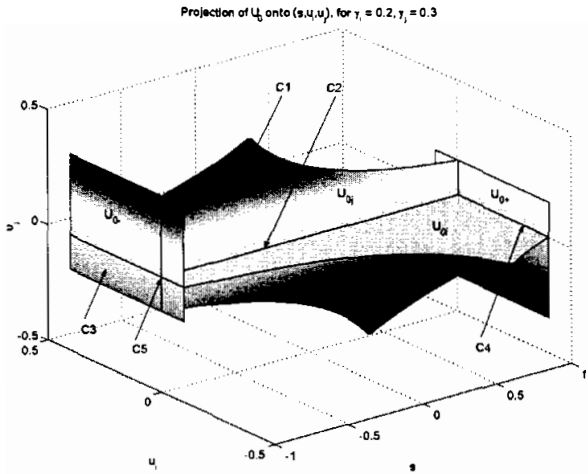

Figure 2: The set $U_0$ projected on $(s, u_i, u_j)$, for $\gamma_i = 0.2$, $\gamma_j = 0.3$. Examples of points of types C1-C5 are marked.

Let $\gamma = (\gamma_1, \ldots, \gamma_n)$ be the model parameters' values of $n$ independent variables that define the $2^n$ dimensional point $Y$, as explicated by the definition of $S'$. The zero set $U_0$ is rather complicated because it contains a number of intersecting multidimensional planes. We have

$$U_0 = \bar{U}_{0-} \cup \bar{U}_{0+} \cup \bigcup_{j=1}^{n} \bar{U}_{0j}$$

where

$$U_{0-} = \left\{(\gamma, u, -1) \mid u_i \in \left(\frac{-\gamma_i}{2}, \frac{1-\gamma_i}{2}\right), i = 1, \ldots, n\right\},$$
$$U_{0+} = \left\{(\gamma, u, 1) \mid u_i \in \left(\frac{\gamma_i-1}{2}, \frac{\gamma_i}{2}\right), i = 1, \ldots, n\right\},$$
$$U_{0j} = \left\{(x, u, s) \mid \begin{array}{l} x = \gamma, u_i = 0, i \neq j, \\ u_j \in \left(-\frac{1}{2}, \frac{1}{2}\right), s \in (-1, 1), \\ -\gamma_j < (1-s)u_j < 1 - \gamma_j \\ \gamma_j - 1 < (1+s)u_j < \gamma_j \end{array}\right\}$$

and $\bar{U}_{0-}$, $\bar{U}_{0+}$, $\bar{U}_{0j}$ denote the closures of $U_{0-}$, $U_{0+}$ and $U_{0j}$.

With the zero set containing $n$-dimensional surfaces $U_{0-}$ and $U_{0+}$ in a $2n+1$ dimensional space (Figure 2), we expect the asymptotic formulae to reflect this fact by the appropriate dimensionality drop of $n - 1$. This indeed happens, but to prove it requires to closely examine the form of $f$ near the different minimum points. This evaluation is complicated by the fact that the zero planes intersect each other, and such cases are not covered by a classic Laplace approximation analysis.

The minimum points of $f$ are divided into five sets according to their location in $U_0$ (Figure 2).

**C1.** $(x', u', s') \in U_{0j} \setminus \bigcup_{i \neq j} U_{0i}$.

**C2.** $(x', u', s') = \bigcap_j U_{0j}$.

**C3.** $(x', u', s') \in U_{0-} \cup U_{0+} \setminus \bigcup_j \bar{U}_{0j}$.

**C4.** $(x', u', s') \in U_{0-} \cup U_{0+} \cup \bar{U}_{0j} \setminus \bigcup_{i \neq j} \bar{U}_{0i}$.

**C5.** $(x', u', s') \in (U_{0-} \cup U_{0+}) \bigcap_j \bar{U}_{0j}$.

Among these cases, $C1$ and $C3$ are almost classical, with $f$ being approximated by a quadratic form in $2n - 1$ and $n + 1$ variables, and the cases $C2$, $C4$ and $C5$ are the most complex, since they correspond to the intersection points of hyper-dimensional planes. We illustrate the treatment of such points for case $C2$.

*Case C2:* $(x', u', s') = \bigcap_j U_{0j}$, i.e., $u = 0$, $s \neq \pm 1$. In this case $z_{0,i} = x'_i$ for all $i$ and $z_{0,I} = 0$ for all other $z$'s. Centering $(x, u, s)$ around the minimum point $(x', u', s')$, we get

$$\tilde{f}(x, u, s) \equiv \sum (z_I - z_{I,0i})^2$$
$$= \sum_l [(x_l + x'_l) - x'_l]^2$$
$$\quad + \sum_{l,k} \left[(1 - (s + s')^2)u_l u_k - 0\right]^2 + \ldots$$
$$= \sum_l x_l^2$$
$$\quad + \sum_{l,k} \left[(1 - s'^2)u_l u_k - (s + 2s')s u_l u_k\right]^2$$
$$\quad + \text{"higher order terms"}$$

So, the principal part of $\tilde{f}$, that bounds $\tilde{f}$ within the multiplicative constant near zero, is given by

$$\tilde{f}(x, u, s) \approx \sum_l x_l^2 + \sum_{l,k} u_l^2 u_k^2.$$

The quadratic form in $x_l$'s contributes an $N^{-n/2}$ factor to the integral $\mathbb{J}[N, Y]$. This can be shown by decomposing the integral and integrating out the $x_l$'s. We are left with the evaluation of the integral

$$\mathbb{J}[N] = \int_{(-\epsilon, +\epsilon)^n} e^{-N \sum u_l^2 u_k^2} du.$$

This is precisely the integral evaluated in Section 3 which was found to be asymptotically equal to $cN^{-\frac{n}{4}}$. Thus the contribution of the neighborhood of $(x', u', s')$ to $\mathbb{J}[N, Y]$ is $cN^{-\frac{3n}{4}}$.

In summary, we have decomposed the proof of Theorem 3 for $Y \in S'$ into five possible cases. We have fully analyzed the second case, using Watanabe's method, showing that the contribution to $\mathbb{J}[N, Y]$ is $cN^{-\frac{3n}{4}}$. The dominating contribution in the cases $C3$, $C4$, and $C5$, are all equal to $cN^{-\frac{n+1}{2}}$ (the proof of this claim is omitted due to space limitations). The dominating contribution in case $C1$ is only $cN^{-\frac{2n-1}{2}}$. Also, the various border points of $U_0$ do not contribute more than the corresponding internal points. Thus, $\ln \mathbb{J}[N, Y] = -\frac{n+1}{2} \ln N + O(1)$. Hence, due to Lemma 4, $\ln \mathbb{I}[N, Y] = N f_Y - \frac{n+1}{2} \ln N + O(1)$, which confirms Theorem 3 for $Y \in S'$. ∎